\documentclass[letterpaper, 10 pt, conference]{ieeeconf}  
\IEEEoverridecommandlockouts                              
\usepackage{amsmath}
\usepackage{amsfonts}
\usepackage{booktabs} 
\usepackage{makecell}
\usepackage{graphicx}
\usepackage{color}
\usepackage{multirow}
\usepackage{pifont}
\usepackage{amsmath}
\usepackage{amssymb}
\usepackage{dsfont}
\definecolor{nbarrier}{RGB}{255, 120, 50}
\definecolor{nbicycle}{RGB}{255, 192, 203}
\definecolor{nbus}{RGB}{255, 255, 0}
\definecolor{ncar}{RGB}{0, 150, 245}
\definecolor{nconstruct}{RGB}{0, 255, 255}
\definecolor{nmotor}{RGB}{200, 180, 0}
\definecolor{npedestrian}{RGB}{255, 0, 0}
\definecolor{ntraffic}{RGB}{255, 240, 150}
\definecolor{ntrailer}{RGB}{135, 60, 0}
\definecolor{ntruck}{RGB}{160, 32, 240}
\definecolor{ndriveable}{RGB}{255, 0, 255}
\definecolor{nother}{RGB}{139, 137, 137}
\definecolor{nsidewalk}{RGB}{75, 0, 75}
\definecolor{nterrain}{RGB}{150, 240, 80}
\definecolor{nmanmade}{RGB}{213, 213, 213}
\definecolor{nvegetation}{RGB}{0, 175, 0}

\title{\LARGE \bf
ST-GS: Vision-Based 3D Semantic Occupancy Prediction with Spatial-Temporal Gaussian Splatting
}

\author{Xiaoyang Yan$^*$, Muleilan Pei$^{*,\dagger}$, and Shaojie Shen
\thanks{\textbf{$^{*}$Equal Contribution}. \textbf{$^{\dagger}$Corresponding Author \& Project Lead}.}
\thanks{This work was supported in part by the Hong Kong Ph.D. Fellowship Scheme, in part by the HKUST Postgraduate Studentship, and in part by the HKUST-DJI Joint Innovation Laboratory.}
\thanks{All authors are with the Department of Electronic and Computer Engineering, The Hong Kong University of Science and Technology, Hong Kong SAR, China. Email: {\texttt{\{xyanaq,mpei,eeshaojie\}@ust.hk}}}
}

\begin{document}

\maketitle
\thispagestyle{empty}
\pagestyle{empty}

\begin{abstract}
3D occupancy prediction is critical for comprehensive scene understanding in vision-centric autonomous driving. Recent advances have explored utilizing 3D semantic Gaussians to model occupancy while reducing computational overhead, but they remain constrained by insufficient multi-view spatial interaction and limited multi-frame temporal consistency. To overcome these issues, in this paper, we propose a novel \underline{S}patial-\underline{T}emporal \underline{G}aussian \underline{S}platting (ST-GS) framework to enhance both spatial and temporal modeling in existing Gaussian-based pipelines. Specifically, we develop a guidance-informed spatial aggregation strategy within a dual-mode attention mechanism to strengthen spatial interaction in Gaussian representations. Furthermore, we introduce a geometry-aware temporal fusion scheme that effectively leverages historical context to improve temporal continuity in scene completion. Extensive experiments on the large-scale nuScenes occupancy prediction benchmark showcase that our proposed approach not only achieves state-of-the-art performance but also delivers markedly better temporal consistency compared to existing Gaussian-based methods.
\end{abstract}


\section{Introduction}
Comprehensive 3D scene understanding is a fundamental requirement for modern autonomous driving systems. 
Vision-based methods have gained increasing attention due to their cost-effectiveness and scalability compared to LiDAR-based approaches \cite{pan2024renderocc}. However, accurately modeling complex and irregular objects in dynamic driving scenes remains challenging for mapless driving \cite{pei2025sept}. Semantic occupancy prediction provides a promising solution by jointly estimating volumetric occupancy and semantic labels of arbitrary-shaped objects in 3D space, thereby improving the reliability and safety of autonomous driving in complex environments \cite{marcuzzi2025sfmocc, pei2025advancing}.

Existing vision-based 3D semantic occupancy prediction methods can be broadly grouped into voxel-based representations, Bird’s-Eye View (BEV) projections, and Gaussian-based scene modeling. Voxel-based approaches \cite{li2023voxformer, jiang2023symphonies} discretize the 3D space into regular grids and predict semantic occupancy for each voxel, but their cubic complexity leads to high memory usage and limited resolution. BEV-oriented methods \cite{li2022bevformer, li2022bevdepth} project image features into a top-down view for efficient reasoning, yet inevitably discard fine-grained vertical structures that are crucial for dense 3D reconstruction. Recently, Gaussian-based representations \cite{GaussianFormer,huang2025gaussianformer2} have emerged as a promising alternative, using 3D Gaussians as compact primitives to capture continuous geometry while maintaining highly efficient rendering.

Despite their efficiency and flexibility, existing Gaussian-based approaches encounter two key challenges: (i) they lack the structured spatial priors inherent to grid-based methods, making spatial interaction across views less effective, and (ii) they struggle to maintain temporal consistency across frames, limiting robustness in dynamic environments. To overcome these issues, we aim to strengthen both multi-view spatial interaction and multi-frame temporal consistency in current Gaussian-based semantic occupancy prediction pipelines.

\begin{figure}[t]
    \centering
    \includegraphics[width=0.48\textwidth]{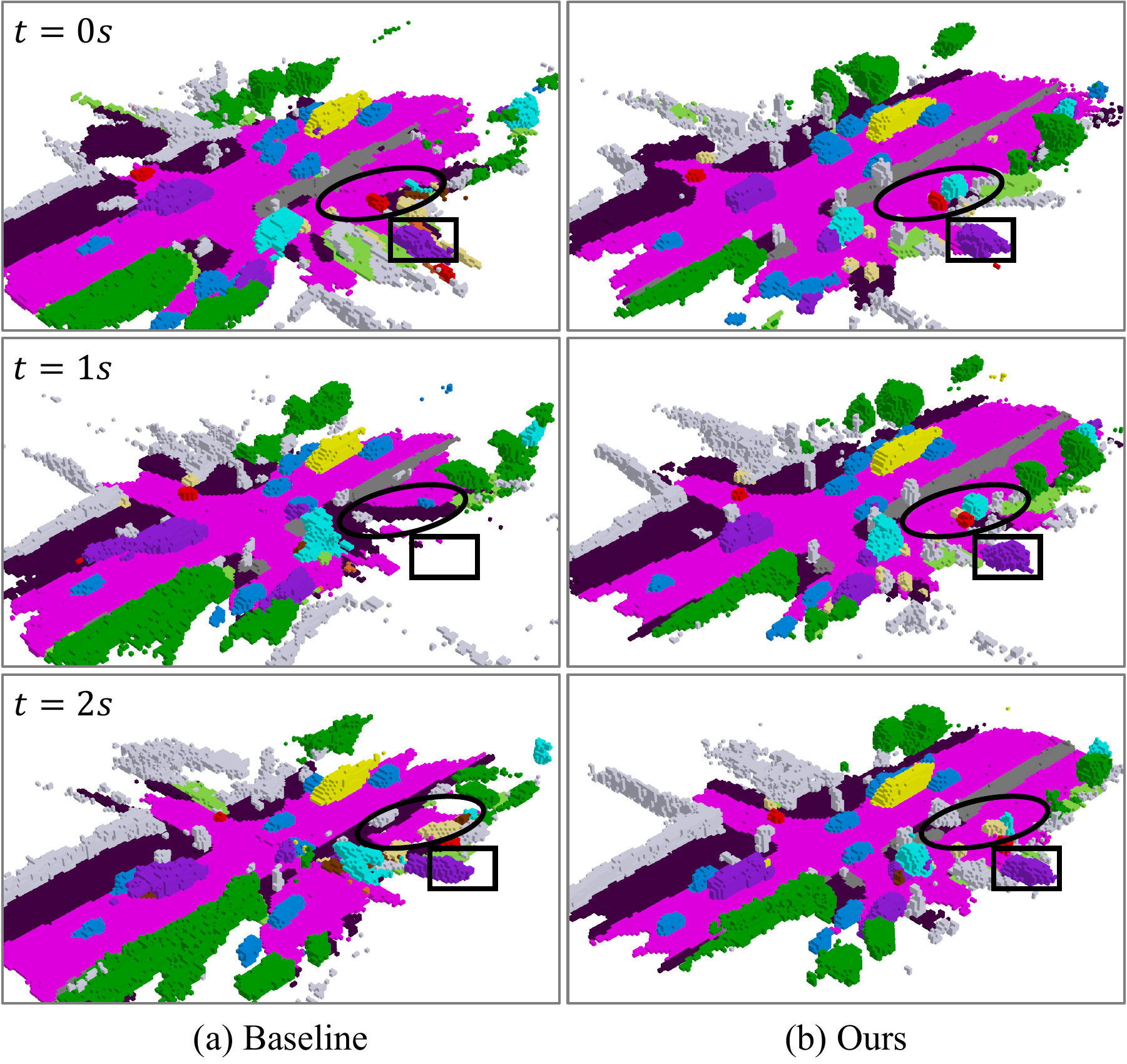}
    \caption{Illustration of temporal inconsistency in occupancy prediction. In this example, the side camera views of the ego vehicle are heavily occluded by surrounding vehicles. The baseline method (GaussianFormer \cite{GaussianFormer}) fails to track the identical truck (highlighted by the box) and produces discontinuous drivable surface predictions (highlighted by the ellipse) across frames. In contrast, our proposed ST-GS effectively integrates historical information, delivering accurate and consistent semantic occupancy predictions.
    }
    \label{fig:fig1}
\end{figure}

Unlike BEV-oriented methods, which employ predefined spatially ordered grid queries that interact with adjacent areas to capture contextual features, 3D Gaussian primitives are spatially independent and lack inherent neighborhood relations. Consequently, Gaussian-based models rely heavily on reference point sampling strategies to extract spatial information from multi-view images. To address this limitation, we introduce a guidance-informed spatial aggregation strategy built on a dual-mode attention mechanism. Specifically, a Gaussian-guided attention module preserves each primitive’s ellipsoidal spatial distribution, while a view-guided attention module aggregates complementary spatial and semantic cues from different perspectives. The reference points of these two attention branches are adaptively fused via an efficient yet effective gated feature aggregation network, producing more robust and spatially aligned Gaussian representations.

Furthermore, as incomplete observations in dynamic driving environments degrade temporal coherence and semantic stability, the existing Gaussian-based method \cite{GaussianFormer} often delivers noticeable temporal inconsistency, as shown in Fig.~\ref{fig:fig1}(a). To mitigate this issue, we propose a geometry-aware temporal fusion scheme that maintains multi-frame Gaussian primitives and incorporates relevant historical information into the current scene representations while explicitly accounting for geometric correspondences. This is achieved by employing a lightweight gated temporal feature fusion module, which significantly enhances temporal consistency across frames in semantic scene completion, as illustrated in Fig.~\ref{fig:fig1}(b).

In summary, our work makes the following contributions:
\begin{itemize}
\item We introduce Spatial-Temporal Gaussian Splatting (ST-GS), a novel framework that effectively improves multi-view spatial interaction and multi-frame temporal consistency for Gaussian-based occupancy prediction.
\item We propose a guidance-informed spatial aggregation strategy within a dual-mode attention mechanism to enhance spatial modeling of 3D Gaussian representations.
\item We design a geometry-aware temporal fusion scheme with a gated feature fusion module to integrate historical contexts while preserving geometric correspondences.
\item Our ST-GS achieves state-of-the-art performance on the nuScenes dataset, and further exhibits superior temporal consistency relative to existing Gaussian-based models.
\end{itemize}


\section{Related Work}
\subsection{3D Semantic Occupancy Prediction}
3D semantic occupancy prediction has gained significant attention in recent years, as it provides a more comprehensive representation of surrounding environments compared to conventional 3D detection or segmentation tasks \cite{pei2021improved}. This capability makes it particularly vital for autonomous driving \cite{pei2025foresight}. Early studies primarily relied on LiDAR point clouds, leveraging their precise geometric information for occupancy estimation \cite{js3cnet, L-CONet}. However, LiDAR sensors are expensive and often degrade under adverse weather or poor lighting conditions. To overcome these limitations, vision-based 3D semantic occupancy prediction has emerged as an active area of research. Recent works have adopted voxel-based representations to model 3D occupancy \cite{wei2023surroundocc, cao2022monoscene, zhang2023occformer, pei2025goirl}, as voxels effectively capture continuous 3D structures within a defined spatial volume. Nonetheless, the inherent sparsity of real-world 3D scenes makes voxel-based methods computationally expensive. Consequently, more efficient scene representations have been explored. TPVFormer \cite{tpvformer} employs a tri-plane representation, which enforces stronger spatial constraints compared to BEV-based approaches \cite{li2022bevformer} that project the scene onto a single plane. Despite these advances, both voxel-based and BEV-based methods require a dense grid of representations to model the 3D environment. In contrast, Gaussian-based approaches encode 3D scenes more compactly, using fewer primitives while maintaining geometric fidelity. Motivated by these advantages, our work focuses on Gaussian-based semantic occupancy prediction.

\subsection{Gaussian-Based Scene Representation}
3D Gaussian representations have been widely adopted in scene reconstruction \cite{kerbl3dgs} due to their strong modeling capabilities and compact encoding of 3D structures. In contrast to dense voxel grids, Gaussian primitives represent 3D semantic occupancy as anisotropic elements that naturally adapt to scene complexity: more primitives are allocated to regions with rich geometry and semantics, while textureless areas are represented with far fewer. Building on these properties, GaussianFormer \cite{GaussianFormer} employs sparse 3D Gaussian primitives and leverages Gaussian-to-voxel splatting to predict semantic occupancy, enabling better adaptability to complicated driving environments than grid-based methods. Furthermore, GaussianFormer-2 \cite{huang2025gaussianformer2} enhances both efficiency and accuracy through introducing a probabilistic distribution modeling strategy to optimize Gaussian initialization, thereby alleviating inefficiencies caused by large empty regions in 3D space. Nevertheless, existing Gaussian-based approaches provide limited treatment of geometric priors and historical information. This motivates our work to strengthen spatial-temporal modeling within Gaussian-based frameworks.

\subsection{Spatial-Temporal Modeling}
Driving scenes are highly dynamic and often suffer from severe occlusions and fast-moving objects, which make reliable 3D semantic occupancy prediction particularly challenging. Several approaches exploit spatial-temporal information to enhance spatial reasoning and improve prediction robustness. PanoOcc \cite{wang2024panoocc} extracts multi-frame image features to construct a set of voxel queries, which are subsequently aligned in unified spatial space and fused through a dedicated temporal encoder to produce a unified occupancy representation. BEVFormer \cite{li2022bevformer} maintains BEV queries across multiple frames and employs a temporal self-attention mechanism to effectively exchange cross-frame features, thereby enhancing scene understanding. GaussianWorld \cite{zuo2024gaussianworld} employs a world-model-driven framework to process video streams to exploit historical observations. ST-Occ \cite{leng2025occupancy} improves the temporal dependency by utilizing spatiotemporal memory across multiple frames. However, current Gaussian-based methods still fall short in spatial-temporal modeling, and thus, we intend to further alleviate this by fully leveraging geometric spatial priors and effectively integrating historical information to boost both prediction accuracy and temporal consistency.


\begin{figure*}[t]
    \centering
    \includegraphics[width=0.98\textwidth]{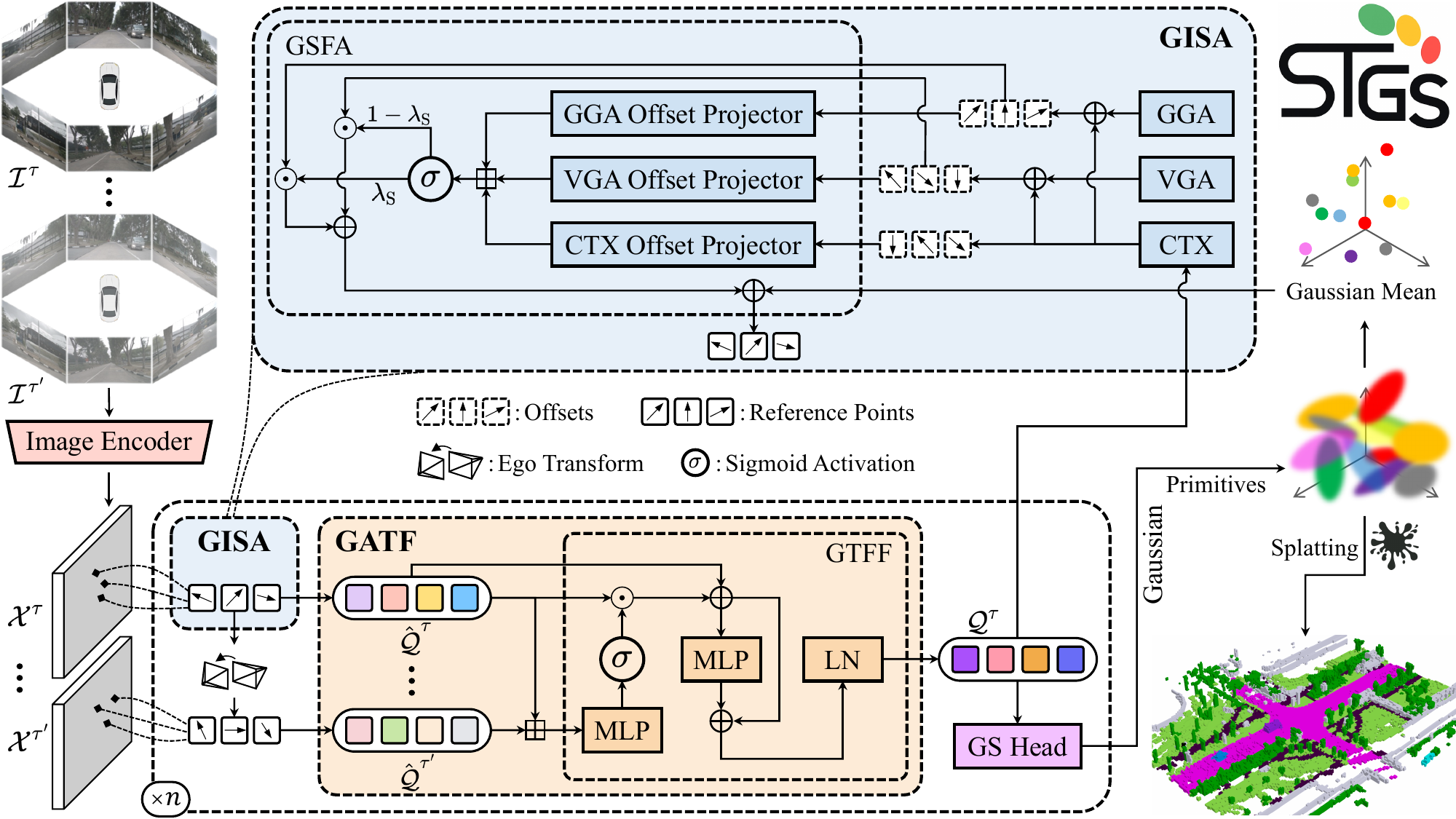}
    \caption{Overview of our ST-GS architecture, demonstrating how it enhances the existing Gaussian-based occupancy prediction model in multi-view spatial interaction and multi-frame temporal consistency.
}
    \label{fig:framework}
\end{figure*}

\section{Methodology}
\subsection{Gaussian-Based Occupancy Prediction}
The 3D semantic occupancy prediction task aims to estimate the volumetric occupancy and semantic labels of each voxel in 3D space. The task takes $N$ images $\mathcal{I} = \{I_{i}\}_{i=1}^{N}$ as input and predicts a dense voxel-based semantic occupancy $\mathcal{O} \in C^{X \times Y \times Z}$, where $C$ is the set of semantic classes. Each voxel is associated with a semantic category. Rather than directly predicting voxel-wise occupancy, the 3D Gaussian representation models a scene as a set of $K$ learnable 3D Gaussian primitives $\mathcal{G} = \{\mathrm{G}_i\}_{i=1}^{K}$. Each primitive $\mathrm{G}_i$ is defined by its center $\mathbf{m}_i \in \mathbb{R}^3$, scale $\mathbf{s}_i \in \mathbb{R}^3$, rotation vectors $\mathbf{r}_i \in \mathbb{R}^4$, opacity $\alpha_i \in \mathbb{R}^1$, and the semantic logits $\mathbf{c}_i \in \mathbb{R}^{|C|}$ corresponding to $|C|$ categories, respectively.
Moreover, we adopt the Gaussian-to-voxel splatting scheme \cite{GaussianFormer} to render Gaussian primitives $\mathcal{G}$ into the voxel space. For each voxel center $\mathbf{x}$, its semantic value can be calculated by:
\begin{equation}
\mathcal{O}(\mathbf{x}) = \sum_{i=1}^{K} \alpha_i \mathrm{exp}(- \frac{1}{2}(\mathbf{x} - \mathbf{m}_i)^{\top} \Sigma^{-1} (\mathbf{x} - \mathbf{m}_i)) \mathbf{c}_i,
\end{equation}
where $\Sigma = RS{S}^{\top}R^{\top}$ denotes the covariance matrix. Herein, $S = \mathrm{diag}(\mathbf{s})$ represents a diagonal matrix whose diagonal entries are the scale components $(s_x, s_y, s_z)$, and $R = \mathrm{q2r}(\mathbf{r})$ denotes the $3 \times 3$ rotation matrix obtained by converting the quaternion $\mathbf{r}$ through the $\mathrm{q2r}(\cdot)$ operation.

\subsection{Framework Overview}
The overall pipeline of our proposed ST-GS framework is illustrated in Fig. \ref{fig:framework}, which enhances the Gaussian-based occupancy prediction paradigm by effectively incorporating spatial-temporal information. Given a sequence of $\tau$ consecutive surround-view images $\{\mathcal{I}^t\}_{t=1}^{\tau}$, we first extract multi-view 2D features $\{\mathcal{X}^t\}_{t=1}^{\tau}$ using a shared image encoder and maintain a set of 3D Gaussian embeddings $\{\mathcal{Q}^t\}_{t=1}^{\tau}$, which act as learnable queries that adaptively sample and aggregate image features to construct 3D representations. Further, the Guidance-Informed Spatial Aggregation (GISA) strategy is introduced to bridge 2D visual features and 3D Gaussian embeddings through a dual-mode attention mechanism: the Gaussian-Guided Attention (GGA) that exploits the intrinsic 3D Gaussian attributes to refine local feature sampling, and the View-Guided Attention (VGA) that leverages spatial-semantic continuity across multi-view images by adaptively sampling along the camera rays. An efficient Gated Spatial Feature Aggregation (GSFA) module is subsequently applied to yield the final reference points. To further improve temporal coherence, the Geometry-Aware Temporal Fusion (GATF) scheme is designed to explicitly align Gaussian embeddings across frames using ego-motion transformations and selectively integrate relevant historical information into the current keyframe representation through an efficient Gated Temporal Feature Fusion (GTFF) module. Finally, the enhanced Gaussian embeddings are decoded into Gaussian primitives by a lightweight GS head \cite{yi2024mvgamba}, which performs Gaussian-to-voxel splatting to generate dense semantic occupancy voxels.

\subsection{Guidance-Informed Spatial Aggregation}
To fully exploit spatial priors from camera viewpoints, we propose the Guidance-Informed Spatial Aggregation (GISA) strategy, which bridges 2D visual features and 3D Gaussian embeddings by dynamically determining how embeddings attend to and query relevant information from the image feature space. GISA employs a dual-mode attention mechanism to incorporate two complementary reference points, enabling more effective spatial feature sampling. Formally, given the 2D image feature maps $\mathcal{X}$ extracted from multi-view cameras and the reference points $\mathcal{P}$ that aggregate offsets from both Gaussian-guided and view-guided attention mechanisms, the single-frame Gaussian embedding $\mathcal{Q} = \{\mathcal{Q}_i\}_{i=1}^K \in \mathbb{R}^{K \times \mathcal{D}}$, with $\mathcal{D}$ denoting the channel dimension, is updated to $\hat{\mathcal{Q}}$ via the following deformable cross-attention operation:
\begin{equation}
\hat{\mathcal{Q}} = \texttt{DeformAttn}(\mathcal{Q}, \mathcal{X}, \mathcal{P}_{2D}),
\label{eq:attention}
\end{equation}
\begin{equation}
\mathcal{P}_{2D} = \texttt{Warp}(\mathcal{P}, \mathcal{K}^\mathrm{cam}, \mathcal{T}^{\mathrm{cam}}),
\end{equation}
where $\texttt{DeformAttn}(\cdot)$ is a deformable cross-attention operation, and $\texttt{Warp}(\cdot)$ denotes the warping function that projects 3D reference points $\mathcal{P}$ onto the 2D image plane, yielding $\mathcal{P}_{2D}$, using the camera intrinsics $\mathcal{K}^\mathrm{cam}$ and extrinsics $\mathcal{T}^{\mathrm{cam}}$.

\begin{figure}[t]
    \centering
    \includegraphics[width=0.486\textwidth]{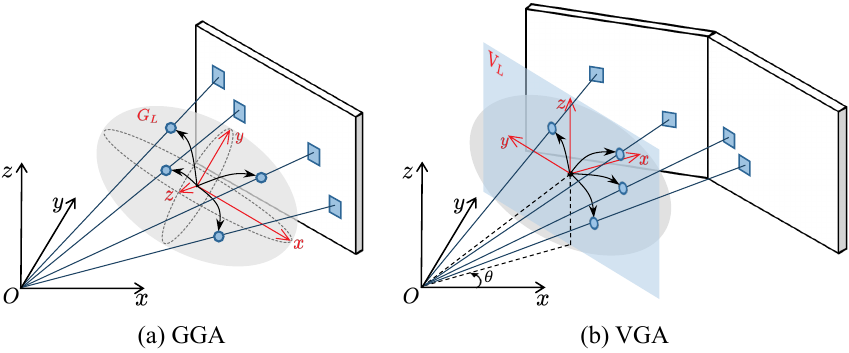}
    \caption{Feature sampling paradigms of offsets for GGA and VGA.}
    \label{fig:GGAVGA}
\end{figure}

\subsubsection{Gaussian-Guided Attention}
The Gaussian-Guided Attention (GGA) mechanism generates adaptive sampling offsets directly from the parameters of each Gaussian, leveraging their intrinsic encoding of the scene’s structural attributes. As shown in Fig. \ref{fig:GGAVGA}(a), GGA uses the Gaussian mean and covariance as geometric guidance to adaptively produce offsets aligned with each Gaussian ellipsoidal distribution.
Further, for each Gaussian instance, we initialize the sampling offsets $\mathcal{P}^{G_L} = \{\mathcal{P}^{G_L}_i \in \mathbb{R}^3\}_{i=1}^M$ consisting of $M$ points structured by 3D grid, with offsets defined in local Gaussian coordinate frame $G_L$. The scaled offset proposal is then combined with learned offsets to form local sampling offsets $\Delta{\mathcal{P}}^{G_L}$:
\begin{equation}
    \Delta \mathcal{P}^{G_L} = s^{G} \mathcal{P}^{G_L} + \Phi_{\Delta}(\mathcal{Q}_i),
    \label{eq:local_pts_g}
\end{equation}  
where $s^G$ is a learnable scaling factor regulating the sampling radius in the Gaussian coordinate space, and $\Phi_{\Delta}(\cdot)$ is the sampling offset predictor.

To transform the local offsets into the perception coordinate frame where each Gaussian is defined, we apply the corresponding rotation $R^G$ and scale $S^G$ transformations to obtain the final 
offset $\Delta \mathcal{P}^G$:
\begin{equation}
\Delta \mathcal{P}^G = R^G S^G  \Delta \mathcal{P}^{G_L}.
\label{eq:offset_g}
\end{equation}

\subsubsection{View-Guided Attention}
Unlike GGA, which utilizes a predefined uniform sampling paradigm consisting of a set of directions evenly distributed in 3D space, motivated by view attention \cite{li2024viewformer}, we design a View-Guided Attention (VGA) mechanism. As demonstrated in Fig. \ref{fig:GGAVGA}(b), VGA generates offsets along the camera viewing directions, enabling more effective spatial information aggregation across overlapping multi-view image features by leveraging cross-view geometric priors. Similarly, for each Gaussian instance, we initialize a set of sampling offsets $\mathcal{P}^{V_L} = \{\mathcal{P}^{V_L}_i \in \mathbb{R}^3\}_{i=1}^M$ based on a 2D grid in the local view coordinate frame $V_L$ (y–z plane). To achieve instance-specific adaptivity, we also predict an offset for each Gaussian instance from its embedding. The local sampling offsets $\Delta{\mathcal{P}}^{V_L}$ can then be obtained by:
\begin{equation}
    \Delta{\mathcal{P}}^{V_L} = s^{V} \mathcal{P}^{V_L} + \Phi_{\Delta}(\mathcal{Q}_i)
    \label{eq:local_pts},
\end{equation}
where $s^V$ is a learnable scalar parameter.
Next, we transform the $V_L$ coordinate frame to the perception coordinate frame by computing a rotation matrix $R^{V}(\theta)$ based on the azimuth angle $\theta$ of the Gaussian center $\mathbf{m}_i$, i.e.,
\begin{equation}
    R^{V}(\theta) = 
    \begin{pmatrix}
        \cos(\theta) & -\sin(\theta) & 0 \\
        \sin(\theta) & \cos(\theta) & 0 \\
        0 & 0 & 1
    \end{pmatrix}
    \label{eq:rot_mat}
\end{equation}
The final VGA offset $\Delta \mathcal{P}^V$ can be obtained by:
\begin{equation}
    \Delta \mathcal{P}^{V} =  R^V(\theta) \Delta \mathcal{P}^{V_L}
    \label{eq:offset_v}.
\end{equation}
 
\subsubsection{Gated Spatial Feature Aggregation}
To better leverage the advantages of both attention mechanisms, we introduce the Gated Spatial Feature Aggregation (GSFA) module to effectively integrate the offsets from GGA and VGA. GSFA employs an attention-driven gating paradigm to dynamically balance the contributions of GGA and VGA.

Given $\Delta \mathcal{P}^G$, $\Delta \mathcal{P}^V$, and the context-aware offset $\Delta \mathcal{P}^{\mathrm{ctx}} = \Phi_{\Delta}(\mathcal{Q}_i)$, we first project them into a latent space, obtaining the corresponding embeddings $\mathcal{F}_G$, $\mathcal{F}_V$, and $\mathcal{F}_{\mathrm{ctx}}$, respectively. These embeddings are then concatenated and passed through a sigmoid activation function $\sigma(\cdot)$ to generate an adaptive gate $\lambda_\mathrm{S} \in [0,1]^{ K \times M}$:  
\begin{equation}
\lambda_{\mathrm{S}} = \sigma(\mathcal{F}_G \boxplus \mathcal{F}_V \boxplus \mathcal{F}_{\mathrm{ctx}}),
\label{eq:gdoa_gate}
\end{equation}
where $\boxplus$ denotes concatenation along the feature dimension.
The final aggregated offset is derived by:
\begin{equation}
\Delta \mathcal{P} = \lambda_\mathrm{S} \odot \Delta \mathcal{P}^G + (1 - \lambda_\mathrm{S}) \odot \Delta \mathcal{P}^V,
\label{eq:gdoa_offset}
\end{equation}
where $\odot$ denotes element-wise multiplication. Consequently, the reference points $\mathcal{P}$ are obtained by adding the aggregated offset $\Delta \mathcal{P}$ to the Gaussian centers $\mathbf{m}_i$:
\begin{equation}
\mathcal{P} = \mathbf{m}_i + \Delta \mathcal{P}.
\label{eq:dual_ref}
\end{equation}

Through GSFA, the dual-mode reference points are seamlessly fused, yielding informative spatial feature attributes.

\subsection{Geometry-Aware Temporal Fusion}
To fully exploit streaming contexts in autonomous driving scenarios, we introduce a Geometry-Aware Temporal Fusion (GATF) scheme that incorporates historical information to enhance the Gaussian representation capability of the current frame. GATF operates on Gaussian embeddings produced by GISA and is designed to model temporal dependencies. By explicitly leveraging ego-motion to establish geometric correspondence across frames and selectively aggregating relevant historical information, GATF significantly enhances multi-frame feature alignment and temporal consistency.

\subsubsection{Inter-frame Geometric Correspondence}
Accurate geometric correspondence between multiple frames is a prerequisite for effective temporal fusion. Since Gaussian embeddings from different timesteps originate from asynchronous observations, the associated reference points are often temporally misaligned. To address this, we explicitly align the reference points of historical frames with those of the current frame, thereby ensuring geometric consistency.

Formally, the reference points $\mathcal{P}^\tau$ of the current frame $\tau$ are transformed into the coordinate system of a historical frame ${\tau^\prime \in [1,\tau-1]}$ through ego-motion information:
\begin{equation}
\mathcal{P}^{\tau^\prime} = \mathcal{T}^{\tau \rightarrow \tau^\prime}  \mathcal{P}^\tau,
\end{equation}
where $\mathcal{P}^{\tau^\prime}$ denotes the aligned reference points in the historical frame, and $\mathcal{T}^{\tau \rightarrow \tau^\prime}$ is the rigid-body transformation from the current frame $\tau$ to the historical frame $\tau^\prime$.

\subsubsection{Gated Temporal Feature Fusion}
Once geometric correspondence is established, the GISA-updated multi-frame Gaussian embeddings $Q = \{ \hat{\mathcal{Q}}^{t} \}_{t=1}^{\tau} \in \mathbb{R}^{ \tau \times K \times \mathcal{D}}$ become available for temporal fusion. The objective is to selectively integrate historical information into the current embedding $\hat{\mathcal{Q}}^{\tau}$, while effectively suppressing inconsistent features arising from occlusions or dynamic objects.

To this end, we introduce a lightweight Gated Temporal Feature Fusion (GTFF) module that adaptively incorporates historical Gaussian embeddings into the current frame. The module first predicts an adaptive fusion gate $\lambda_\mathrm{T} \in [0,1]^{K \times \mathcal{D}}$ via a temporal weight generator:
\begin{equation}
\lambda_\mathrm{T}  = \sigma \big( \texttt{MLP}(Q) \big),
\end{equation}
where $\texttt{MLP}(\cdot)$ represents a multi-layer perceptron block.

Next, the gate $\lambda_\mathrm{T} $ modulates the contribution of historical embeddings relative to the current frame, producing a gated embedding $\tilde{\mathcal{Q}}^{\tau}$:
\begin{equation}
    \tilde{\mathcal{Q}}^{\tau} = \hat{\mathcal{Q}}^{\tau} + \lambda_\mathrm{T}  \odot \hat{\mathcal{Q}}^{\tau}.
\end{equation}

Consequently, the final Gaussian embedding of the current frame, $\mathcal{Q}^{\tau} \in \mathbb{R}^{K \times \mathcal{D}}$, is obtained via a residual refinement:
\begin{equation}
    \mathcal{Q}^{\tau} = \texttt{LN}\Big(\hat{\mathcal{Q}}^{\tau} + \texttt{MLP}(\tilde{\mathcal{Q}}^{\tau})\Big),
\end{equation}
where $\texttt{LN}(\cdot)$ denotes layer normalization.

By jointly learning adaptive temporal weights and residual refinement, the GTFF module effectively integrates historical information, strengthening multi-frame temporal consistency.

\subsection{Training Loss}
After the Gaussian-to-voxel splatting process, we obtain the semantic occupancy prediction. Consistent with \cite{GaussianFormer, tpvformer}, we adopt the cross entropy loss and the Lov\'{a}sz-Softmax loss \cite{berman2018lovaszloss} to optimize the output of each block.


\begin{table*}[t]
    \caption{3D semantic occupancy prediction results on the nuScenes validation split.} 
    \small
    \setlength{\tabcolsep}{0.005\linewidth}  
    \vspace{-1.5mm}  
    \renewcommand\arraystretch{1.05}
    \centering
    \resizebox{\textwidth}{!}{
    \begin{tabular}{l| r | c c | c c c c c c c c c c c c c c c c}
        \toprule
        Method & \multicolumn{1}{c|}{Venue} 
        & \makecell{SC \\ IoU} & \makecell{SSC \\ mIoU}
        & \rotatebox{90}{\textcolor{nbarrier}{$\blacksquare$} barrier}
        & \rotatebox{90}{\textcolor{nbicycle}{$\blacksquare$} bicycle}
        & \rotatebox{90}{\textcolor{nbus}{$\blacksquare$} bus}
        & \rotatebox{90}{\textcolor{ncar}{$\blacksquare$} car}
        & \rotatebox{90}{\textcolor{nconstruct}{$\blacksquare$} const. veh.}
        & \rotatebox{90}{\textcolor{nmotor}{$\blacksquare$} motorcycle}
        & \rotatebox{90}{\textcolor{npedestrian}{$\blacksquare$} pedestrian}
        & \rotatebox{90}{\textcolor{ntraffic}{$\blacksquare$} traffic cone}
        & \rotatebox{90}{\textcolor{ntrailer}{$\blacksquare$} trailer}
        & \rotatebox{90}{\textcolor{ntruck}{$\blacksquare$} truck}
        & \rotatebox{90}{\textcolor{ndriveable}{$\blacksquare$} drive. suf.}
        & \rotatebox{90}{\textcolor{nother}{$\blacksquare$} other flat}
        & \rotatebox{90}{\textcolor{nsidewalk}{$\blacksquare$} sidewalk}
        & \rotatebox{90}{\textcolor{nterrain}{$\blacksquare$} terrain}
        & \rotatebox{90}{\textcolor{nmanmade}{$\blacksquare$} manmade}
        & \rotatebox{90}{\textcolor{nvegetation}{$\blacksquare$} vegetation} \\
        \midrule
        MonoScene~\cite{cao2022monoscene} & CVPR 2022 & 23.96 & 7.31 & 4.03 & 0.35& 8.00& 8.04&	2.90& 0.28& 1.16&	0.67&	4.01& 4.35&	27.72&	5.20& 15.13&	11.29&	9.03&	14.86 \\
        Atlas~\cite{murez2020atlas} & ECCV 2020 & 28.66 & 15.00 & 10.64 &5.68 &19.66& 24.94& 8.90&	8.84&	6.47& 3.28&	10.42&	16.21&	34.86&	15.46&	21.89&	20.95&	11.21&	20.54 \\
        BEVFormer~\cite{li2022bevformer} & ECCV 2022 & 30.50 & 16.75 & 14.22 &	6.58 & 23.46 & 28.28& 8.66 &10.77& 6.64& 4.05& 11.20&	17.78 & 37.28 & 18.00 & 22.88 & 22.17 & 13.80 &	\underline{22.21}\\
        TPVFormer~\cite{tpvformer} & CVPR 2023 & 11.51 & 11.66 & 16.14&	7.17& 22.63	& 17.13 & 8.83 & 11.39 & 10.46 & 8.23&	9.43 & 17.02 & 8.07 & 13.64 & 13.85 & 10.34 & 4.90 & 7.37\\
        TPVFormer$^\dagger$\cite{tpvformer}  & CVPR 2023 & {30.86} & 17.10 & 15.96&	 5.31& 23.86 & 27.32 & 9.79 & 8.74 & 7.09 & 5.20& 10.97 & 19.22 & {38.87} & {21.25} & {24.26} & {23.15} & 11.73 & 20.81\\
        OccFormer~\cite{zhang2023occformer} & ICCV 2023 & {31.39} & {19.03} & {18.65} & {10.41} & {23.92} & {30.29} & {10.31} & {14.19} & {13.59} & {10.13} & {12.49} & {20.77} & {38.78} & 19.79 & 24.19 & 22.21 & {13.48} & {21.35}\\
        SurroundOcc~\cite{wei2023surroundocc} & ICCV 2023 & \underline{31.49} & \underline{20.30}  & \underline{20.59} & {11.68} & \textbf{28.06} & \underline{30.86} & {10.70} & {15.14} & {\textbf{14.09}} & {\textbf{12.06}} & \underline{14.38} & \underline{22.26} & 37.29 & \underline{23.70} & {24.49} & {22.77} & {\underline{14.89}} & {21.86}  \\
        GaussianFormer~\cite{GaussianFormer} & ECCV 2024 & 29.83 & {19.10} & {19.52} & {11.26} & {26.11} & {29.78} & {10.47} & {13.83} & {12.58} & {8.67} & {12.74} & {21.57} & {39.63} & {23.28} & {24.46} & {22.99} & 9.59 & 19.12 \\
        GaussianFormer-2*\cite{huang2025gaussianformer2} & CVPR 2025 & 30.56 & 20.02 & 20.15 & \underline{12.99} & 27.61 & 30.23 & \underline{11.19} & \underline{15.31} & 12.64 & 9.63 & 13.31 & \underline{22.26} & \underline{39.68} & 23.47 & \underline{25.62} & \underline{23.20} & 12.25 & 20.73 \\
        \midrule
        \textbf{ST-GS(Ours)} & ICRA 2026 & \textbf{32.88} & \textbf{21.43} & \textbf{21.04} & \textbf{14.13} & \underline{27.78} & \textbf{31.62} & \textbf{11.85} & \textbf{17.84} & \underline{13.63} & \underline{10.76} & \textbf{14.85} & \textbf{23.22} & \textbf{41.88} & \textbf{24.40} & \textbf{26.71} & \textbf{24.70} & \textbf{15.00} & \textbf{23.48}  \\
        \bottomrule
    \end{tabular}}
    \begin{flushleft}{~~$\dagger$ means supervision with dense occupancy annotations \cite{wei2023surroundocc}. \\
    ~~* indicates results of the 128-channel dimension for a fair comparison \cite{huang2025gaussianformer2}.}\end{flushleft}
    \label{tab: nuscenes results}
    \vspace{-0.45cm}
\end{table*}

\section{Experiments and Results}
\subsection{Experimental Setup}
\subsubsection{Dataset}
We evaluate our approach on the nuScenes dataset \cite{nuscenes}, which consists of 1,000 driving sequences, each lasting 20 seconds. The dataset provides annotations at 2 Hz and includes six synchronized cameras covering a full $360^{\circ}$ horizontal field of view. Following the established protocol \cite{GaussianFormer, huang2025gaussianformer2}, we adopt the official split of 700 scenes for training and 150 for validation. For supervision, we use the semantic occupancy ground truth provided by SurroundOcc \cite{wei2023surroundocc}. The annotated space spans a volume of $100m \times 100m \times 8m$, centered on the ego vehicle. The target occupancy representation is discretized into a $200 \times 200 \times 16$ voxel grid, where each voxel corresponds to $0.5m$ in physical space. This covers spatial ranges of $[-50m, 50m]$ along both horizontal axes and $[-5m, 3m]$ in height. RGB images from all six cameras are used at their native resolution of $1600 \times 900$ pixels.

\subsubsection{Evaluation Metrics}
We assess the performance of 3D semantic occupancy prediction using two standard metrics following \cite{cao2022monoscene}. For Scene Completion (SC), we report class-agnostic Intersection-over-Union (IoU), which evaluates geometric accuracy irrespective of semantics. For Semantic Scene Completion (SSC), we report mean IoU (mIoU) across all semantic classes, reflecting both geometric and semantic quality. Formally,
\begin{equation}
\mathrm{IoU} = \frac{\mathrm{TP}^{(c_n)}}{\mathrm{TP}^{(c_n)} + \mathrm{FP}^{(c_n)} + \mathrm{FN}^{(c_n)}},
\end{equation}
\begin{equation}
\mathrm{mIoU} = \frac{1}{|C^\prime|} \sum_{i \in C^\prime} \frac{\mathrm{TP}^{(i)}}{\mathrm{TP}^{(i)} + \mathrm{FP}^{(i)} + \mathrm{FN}^{(i)}},
\end{equation}
where $\mathrm{TP}^{(i)}$, $\mathrm{FP}^{(i)}$, and $\mathrm{FN}^{(i)}$ denote the true positives, false positives, and false negatives, respectively, for class $i \in C^\prime$, and $C^\prime$ is the set of non-empty semantic classes. The variable $c_n$ refers to non-empty classes in the SC evaluation.

Moreover, we employ the Spatial-Temporal Classification Variability (STCV) metric \cite{leng2025occupancy} to quantify the temporal consistency of occupancy predictions across consecutive frames. STCV calculates the proportion of classification alterations in non-empty voxels relative to all non-empty voxels over $\mathrm{L}$ successive frames within each scene, defined as:
\begin{equation}
    \mathrm{STCV} = 
    \frac{1}{\mathrm{L-1}} \sum_{t=1}^{\mathrm{L-1}} 
    \frac{
        \sum_{\mathcal{V}_t \land  \mathcal{V}_{t+1}} \mathds{1}\!\left[
            \mathcal{O}^{t} \neq \mathcal{O}^{t+1} 
        \right]
    }{
        \sum_{\mathcal{V}_t \land  \mathcal{V}_{t+1}} \mathds{1}\!\left[
            \mathcal{O}^{t}
        \right]
    },
\end{equation}
where $\mathcal{V}_t$ denotes the set of non-empty voxels at the frame $t$, and $\mathds{1}[\cdot]$ represents the indicator function. For a comprehensive evaluation of temporal consistency, we report the mean STCV (mSTCV), minimum STCV (minSTCV), and maximum STCV (maxSTCV) across all scenes.

\subsubsection{Implementation Details}
We adopt the default ResNet-101-DCN \cite{resnet} as the backbone for image feature extraction, consistent with existing Gaussian-based methods \cite{GaussianFormer, huang2025gaussianformer2}, to ensure fair comparison. A Feature Pyramid Network (FPN) \cite{fpn} is employed to capture multi-scale features at downsampling ratios of $\{4\times, 8\times, 16\times, 32\times\}$. The channel dimension $\mathcal{D}$ is fixed to 128 across all models. The Gaussian-based decoder comprises $n=4$ stacked blocks with $K=25,600$ Gaussian primitives. We train our model using AdamW~\cite{adamw} with a weight decay of 0.01. The learning rate follows a warm-up strategy for the first 500 iterations, reaching a maximum value of $2 \times 10^{-4}$, and then decays following a cosine annealing schedule. To improve model generalization, we leverage standard data augmentations, including random cropping, flipping, resizing, and photometric distortions.

\subsection{Quantitative Results}
\subsubsection{Main Results}
Table \ref{tab: nuscenes results} offers a comprehensive comparison of our ST-GS with existing methods on the nuScenes validation split, with the best and second-best results highlighted in \textbf{bold} and \underline{underlined}, respectively. Our ST-GS consistently outperforms both previous voxel-based approaches and recent Gaussian-based methods. Compared to our baseline, GaussianFormer \cite{GaussianFormer}, ST-GS achieves notable improvements of 10.22\% in IoU and 12.20\% in mIoU. Even against its successor, GaussianFormer-2 \cite{huang2025gaussianformer2}, ST-GS delivers substantial gains of 7.59\% in IoU and 7.04\% in mIoU. Moreover, our method attains the highest performance across most semantic categories. These results demonstrate the effectiveness of our framework and underscore that enhancing spatial interactions and incorporating temporal information are both crucial for accurate 3D semantic occupancy prediction.

\begin{table}[t]
\centering
\caption{Evaluation of temporal consistency on the nuScenes dataset}
\setlength{\tabcolsep}{1.2mm}
\begin{tabular}{l|ccc}
\toprule
Method  & mSTCV(\%) & minSTCV(\%) & maxSTCV(\%)\\
\midrule
FB-OCC \cite{fbocc} & 12.18 & - & - \\
ST-Occ \cite{leng2025occupancy} & 8.68 & - & - \\
GaussianFormer \cite{GaussianFormer} & 6.52 & 1.59 & 12.71 \\
GaussianFormer-2 \cite{huang2025gaussianformer2} & 5.97 & 1.94 & 10.86 \\
\midrule
\textbf{ST-GS (Ours)} & \textbf{4.47} & \textbf{1.03} & \textbf{8.53} \\
\bottomrule
\end{tabular}
\begin{flushleft}{~~~~A ``-" denotes the lack of relevant data.}\end{flushleft}
\vspace{-0.25cm}
\label{tab:temporal consistency}
\end{table}

\subsubsection{Temporal Consistency Results}
We conduct the evaluation of temporal consistency on the nuScenes validation set. All metrics follow the lower-is-better criterion. As shown in Table \ref{tab:temporal consistency}, our ST-GS achieves the best performance across all metrics, surpassing both previous voxel-based approaches and recent Gaussian-based methods. In particular, relative to our baseline, GaussianFormer \cite{GaussianFormer}, ST-GS reduces temporal inconsistency by 31.44\% in mSTCV, 35.22\% in minSTCV, and 32.89\% in maxSTCV, respectively. These results highlight that our framework substantially improves multi-frame temporal consistency, leading to more robust and stable 3D semantic occupancy prediction.

\begin{figure*}[t]
    \centering
    \includegraphics[width=0.98\textwidth]{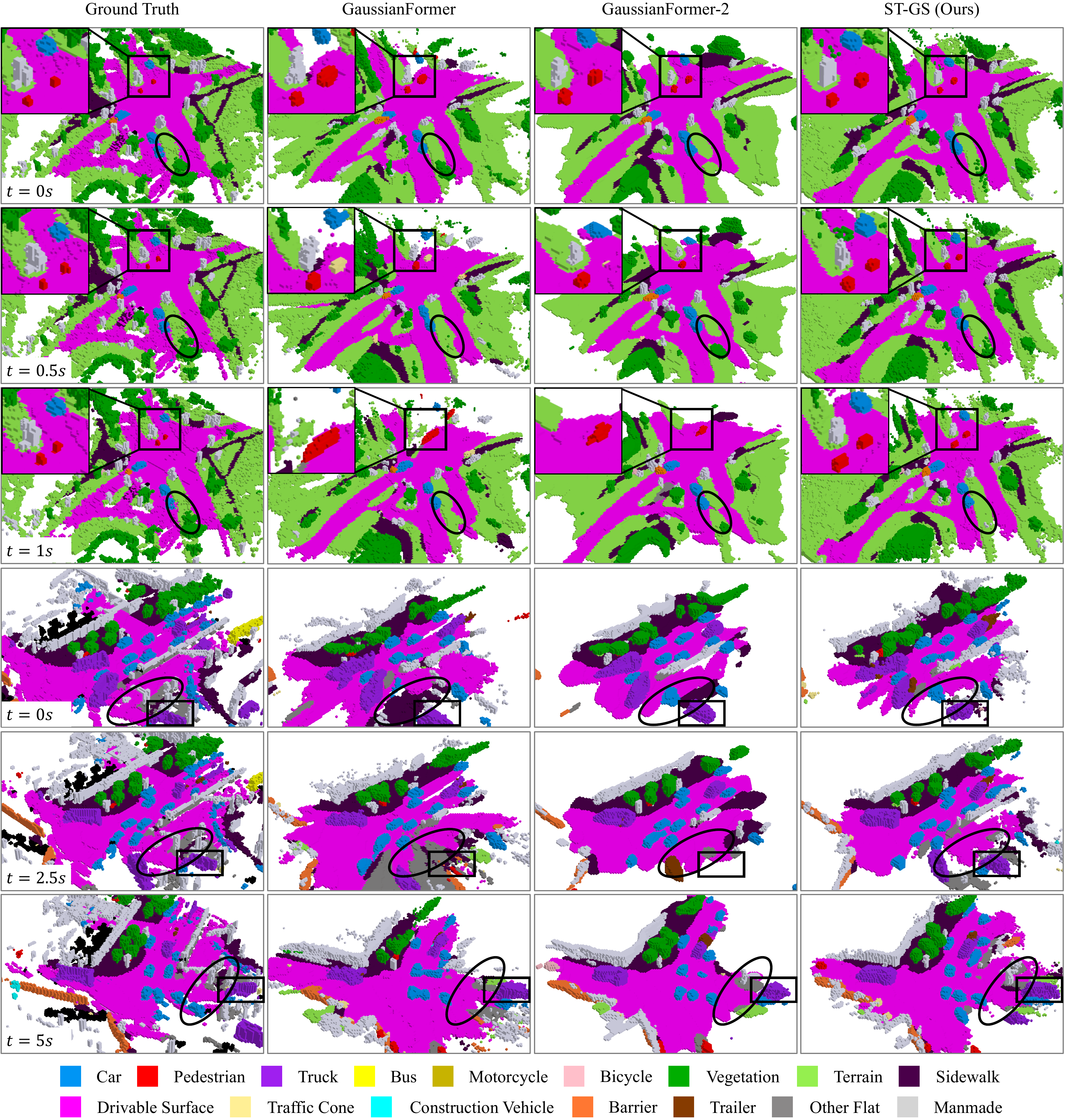}
    \caption{Qualitative comparison of the baseline GaussianFormer \cite{GaussianFormer}, GaussianFormer-2 \cite{huang2025gaussianformer2}, and our proposed ST-GS. Visualization results of three-timestamp predictions from two distinct driving sequences show that ST-GS delivers more spatially accurate and temporally consistent semantic occupancy predictions.}
    \label{fig:visualization}
    \vspace{-0.2cm}
\end{figure*}

\subsection{Ablation Studies}
We perform ablation studies on the nuScenes validation split to assess the effectiveness of the proposed GISA strategy and GATF scheme.

\subsubsection{Effect of the GISA Strategy}
In Table \ref{tab:GISA}, we present the ablation results for each component of GISA, including GGA, VGA, and GSFA. Compared with the baseline GaussianFormer \cite{GaussianFormer}, both GGA and VGA individually yield improvements in IoU and mIoU, validating the contribution of each module. Moreover, when combined through GSFA, performance is further enhanced, highlighting the advantage of the fusion strategy in facilitating multi-view spatial feature interactions and achieving substantial performance gains in 3D semantic occupancy prediction accuracy.

\begin{table}[tbp]
    \centering
    \caption{Ablation on components of the GISA Strategy.}
    \label{tab:GISA}
    \setlength{\tabcolsep}{1.2mm}
    \begin{tabular}{l|ccc|cc}
        \toprule
       Method & GGA & VGA & GSFA & IoU & mIoU \\
        \midrule    
         GaussianFormer~\cite{GaussianFormer} &&&& 29.83  & 19.10 \\
        w/ GGA Only & \ding{51} &&& 30.91 & 19.85 \\
        w/ VGA Only && \ding{51} && 30.97 & 19.92 \\
        w/ GISA (Ours) & \ding{51} & \ding{51} & \ding{51} & \textbf{31.51} & \textbf{20.27} \\
        \bottomrule
    \end{tabular}
\end{table}

\begin{table}[tbp]
    \begin{minipage}[t]{0.24\textwidth}
    \centering
    \makeatletter\def\@captype{table}\makeatother\caption{Effect of sequence length.}
    \label{tab:temporal sequence length}
    \setlength{\tabcolsep}{1.2mm}
    \begin{tabular}{c|cc}
        \toprule
        \# of frames & IoU & mIoU \\
        \midrule    
         1 & 31.51 & 20.27  \\
         2 & 32.01 & 20.82 \\
         3 & \textbf{32.88} & \textbf{21.43} \\
        \bottomrule
    \end{tabular}
    \end{minipage}
    \begin{minipage}[t]{0.24\textwidth}
    \centering
    \makeatletter\def\@captype{table}\makeatother\caption{Effect of fusion mode.}
    \label{tab:fusion strategies}
    \setlength{\tabcolsep}{1.2mm}
    \begin{tabular}{l|cc}
        \toprule
        Fusion Mode & IoU & mIoU \\
        \midrule    
         Loose & 32.11 & 21.11 \\
         Tight & 32.44 & 21.13 \\
         Coupled & \textbf{32.88} & \textbf{21.43} \\
        \bottomrule
    \end{tabular}
    \end{minipage}
\end{table}

\subsubsection{Effect of the GATF Scheme}
We further evaluate the effectiveness of the proposed GATF scheme. As shown in Table \ref{tab:temporal sequence length}, we first investigate the effect of temporal sequence length. The results demonstrate that increasing the number of frames consistently improves all metrics, confirming that appropriately integrating richer historical context enhances prediction performance, as long as computational overhead remains controlled.
We then analyze the impact of different fusion modes of GTFF on prediction accuracy. Specifically, we examine three distinct fusion configurations: (i) loose mode, where GTFF aggregates historical embeddings into the current frame only once at the final stage after the four stacked blocks; (ii) tight mode, where GTFF is applied within each individual block for fine-grained temporal fusion; and (iii) coupled mode, which combines both strategies to attain more comprehensive feature integration. As reported in Table \ref{tab:fusion strategies}, all fusion modes enhance prediction performance, with the coupled mode delivering the largest improvements in both IoU and mIoU. These findings underscore that incorporating richer temporal information and adopting effective fusion schemes are essential for achieving significant performance gains in 3D semantic occupancy prediction.

\subsection{Qualitative Results}
We present visualizations of two representative scenarios with varying time intervals from the nuScenes validation set to demonstrate the superiority of our ST-GS over the baseline GaussianFormer \cite{GaussianFormer} and GaussianFormer-2 \cite{huang2025gaussianformer2}. As shown in the upper group of Fig. \ref{fig:visualization}, all methods correctly predict two pedestrians and one car (box) in the initial frame, whereas GaussianFormer-2 misclassifies the drivable surface (ellipse). In the subsequent frame, GaussianFormer misclassifies one pedestrian category, and GaussianFormer-2 fails to detect one pedestrian. In the final frame, both fail to accurately predict the pedestrians and the car. In contrast, our ST-GS preserves category accuracy and instance-level stability across all three consecutive frames, effectively handling both small objects and large structures. Furthermore, the lower group of Fig. \ref{fig:visualization} showcases an intersection with heavy occlusion over a longer time span. In this case, both of the other two methods fail to stably track the same truck (box) and generate temporally inconsistent predictions for drivable surfaces (ellipse), whereas our ST-GS accurately and consistently identifies them across frames. Overall, these qualitative results highlight that ST-GS substantially improves both prediction accuracy and temporal consistency, even under long-horizon scenarios. Additional qualitative results are provided in the supplementary video.


\section{Conclusion}
In this paper, we propose ST-GS, an innovative framework designed to strengthen both spatial and temporal modeling in the Gaussian-based semantic occupancy prediction pipeline. Specifically, ST-GS improves multi-view spatial interaction through the GISA strategy and enforces multi-frame temporal consistency via the GATF scheme. Experimental results on the large-scale nuScenes occupancy prediction benchmark exhibit that ST-GS achieves 32.88 IoU and 21.43 mIoU, outperforming prior voxel-based methods and recent Gaussian-based approaches by substantial margins. Furthermore, ST-GS reduces temporal inconsistency by over 30\% in mSTCV relative to the baseline, demonstrating the effectiveness of enhancing spatial interactions and incorporating temporal information in delivering accurate and robust 3D semantic occupancy predictions. Future work will focus on improving the efficiency of the proposed framework by exploring advanced architectures, such as Gamba \cite{shen2024gamba} which integrates Gaussian splatting with the Mamba \cite{mamba} training paradigm.


\bibliography{reference}
\end{document}